\documentclass{article}

\usepackage{arxiv}
\usepackage{listings}
\usepackage{array}
\usepackage{xcolor}
\usepackage[utf8]{inputenc} % allow utf-8 input
\usepackage[T1]{fontenc}    % use 8-bit T1 fonts
\usepackage{hyperref}       % hyperlinks
\usepackage{url}            % simple URL typesetting
\usepackage{booktabs}       % professional-quality tables
\usepackage{amsfonts}       % blackboard math symbols
\usepackage{nicefrac}       % compact symbols for 1/2, etc.
\usepackage{microtype}      % microtypography
\usepackage{lipsum}		% Can be removed after putting your text content
\usepackage{graphicx}
\usepackage{doi}

\title{Increasing the LLM Accuracy for Question Answering: Ontologies to the Rescue!}

\author{ Dean Allemang \\
	data.world AI Lab \\
	\texttt{dean.allemang@data.world} \\
 \And
	 Juan F. Sequeda\\
	data.world AI Lab \\
	\texttt{juan@data.world} \\
}

\renewcommand{\headeright}{Technical Report}
\renewcommand{\undertitle}{Technical Report}

\renewcommand{\shorttitle}{Increasing the Accuracy of LLM QA Systems with Ontologies}

\hypersetup{
pdftitle={Increasing the Accuracy of LLM Question-Answering Systems with Ontologies},
pdfauthor={Dean Allemang and Juan Sequeda, },
}

\begin{document}
\maketitle

\begin{abstract}
There is increasing evidence that question-answering (QA) systems with Large Language Models (LLMs), which employ a knowledge graph/semantic representation of an enterprise SQL database (i.e. Text-to-SPARQL), achieve higher accuracy compared to systems that answer questions directly on SQL databases (i.e. Text-to-SQL). 
Our previous benchmark research showed that by using a knowledge graph, the accuracy improved from 16\% to 54\%.
The question remains: how can we further improve the accuracy and reduce the error rate?
Building on the observations of our previous research where the inaccurate LLM-generated SPARQL queries followed incorrect paths, we present an approach that consists of 1) \textbf{Ontology-based Query Check (OBQC)}: detects errors by leveraging the ontology of the knowledge graph to check if the LLM-generated SPARQL query matches the semantic of ontology and 2) \textbf{LLM Repair}: use the error explanations with an LLM to repair the SPARQL query. 
Using the chat with the data benchmark, our primary finding is that our approach increases the overall accuracy to 72\% including an additional 8\% of \textit{``I don't know"} unknown results. 
Thus, the overall error rate is 20\%.
% There is increasing evidence that question-answering (QA) systems with Large Language Models (LLMs), which employ a knowledge graph representation of an enterprise SQL database (Text-to-SPARQL), achieve higher accuracy compared to systems that answer questions directly on SQL databases (Text-to-SQL). 
% The objective of this research is to further improve the accuracy of these LLM Question Answering systems. 
% Our approach, Ontology-based Query Check (OBQC), is to check the LLM generated SPARQL query against the semantics specified by the ontology. 
% A query will be flagged as incorrect and prevented from execution if it does not align with the ontological semantics. 
% The study also explores the LLM’s capability in repairing a SPARQL query given an explanation of the error (LLM Repair). 
% Our methods are evaluated using the chat with the data benchmark. 
% The primary finding is our method further increases the accuracy overall by  21.59\% thus pushing the overall accuracy level to 65.63\%. 
These results provide further evidence that investing knowledge graphs, namely the ontology, provides higher accuracy for LLM powered question answering systems.
%Our method is a component of the data.world AI Context Engine which is being widely used by customers in Generative AI production use cases that enable business users to chat with SQL databases.

\end{abstract}

% keywords can be removed

\section{Introduction}

Business users and executives would like to have an expert about their business available to them at all times in order to ask questions and receive accurate, explainable and governed answers. 
In order to achieve this goal, there has been decades of research on question answering on structured data, namely SQL databases \cite{10.1145/1460690.1460714,10.1145/800186.810578,10.1145/355598.362773,10.1145/320251.320253,data-atis-original,data-geography-original,data-restaurants-logic}. 
With the rise of Generative AI and Large Language Models, this desire is stronger than ever. 
We postulate that LLMs combined with Knowledge Graphs are enablers of this vision.
Our previous research provided evidence that an LLM powered question answering system that answers enterprise natural language questions over a Knowledge Graph representation of an enterprise SQL database returns 3x more accurate results than a LLM without a Knowledge Graph \cite{OurPreviousWork}. 

Enterprises will require higher accuracy in order to adopt these question answering systems. 
Thus there is a critical need to explore deterministic approaches to increase the accuracy. 
Without improved accuracy, organizations risk a lack of adoption of question answering powered LLM systems, thus not fulfilling the vision.

Based on our previous research \cite{OurPreviousWork}, our intuition is that accuracy can be further increased by 1) leveraging the ontology of the knowledge graph to check for errors in the queries and 2) using the LLM to repair incorrect queries.

First, we observed identifiable patterns to how generated queries would fail. 
One such pattern was that the generated query would use a property in the wrong direction; 
for example in the insurance domain, \textit{a Policy is sold by an Agent} is the accurate knowledge that would be represented in an ontology. 
However, if a query would represent \textit{an Agent is sold by a Policy}, it would not match the semantics in the ontology because that is not what is represented in the ontology. 
Our intuition is that the semantics of the ontology can detect these types of errors.

For example, assume the following question \textit{``return all the policies that an agent sold"}, resulted in the following SPARQL query:

%\begin{lstlisting}[language=SPARQL]
\begin{verbatim}
SELECT ?agent ?policy
WHERE {
  ?agent :soldByAgent ?policy . 
  ?agent rdf:type :Agent 
}
\end{verbatim}
%\end{lstlisting}

and given the following snippet of an ontology

\begin{verbatim}
:soldByAgent 
  rdfs:domain :Policy ;
  rdfs:range :Agent .
\end{verbatim}

we could determine that the generated query should be correct correct if the domain of \texttt{:soldByAgent} is \texttt{:Policy}.
However, per the query, the domain is \texttt{:Agent} and assuming they are disjoint, the generated query is incorrect. 

This first part of our approach is the \textbf{Ontology-based Query Check} (OBQC), which checks if the query is valid by applying rules based on the semantics of the ontology. 
A set of rules check the the body of the query (i.e. the WHERE clause).
Another set of rules checks the head of query (i.e. the SELECT clause). 
It is important to clarify that the Ontology-based Query Check does not use an LLM. 
It is deterministic and based solely on the semantics of the ontology. 
If the OBQC detects an error, we could either determine to not return the result thus terminate or we could try to repair the query.
This leads us to our second intuition: repairing.

Second, repairing databases\cite{10.1145/1514894.1514899} and programs\cite{10.1145/3318162,10.1145/3631974} has been an long standing research area in computer science. 
Recently, LLMs have been applied to repair programs\cite{bouzenia2024repairagent, 10.1145/3611643.3613892}.
Furthermore, new approaches are arising to correct the results coming from LLMs through retrieval augmented generation\cite{yan2024corrective}.
Inspired by these approaches, consider the following example. 
Once a SPARQL query is detected to be incorrect, we can define an explanation for the reason why it is correct. 
Per our running example, an explanation is the following: \textit{The property
:soldByAgent has domain :Policy, but its subject ?agent is a :Agent, which isn’t a subclass of :Policy.}.
What if we can pass the incorrect SPARQL query, with this explanation and prompt the LLM to rewrite the query? 

The second part of our approach is \textbf{LLM Repair}, which repairs the SPARQL query generated by the LLM. 
It takes as input the incorrect query and the explanation coming from the rule(s) that was fired as an explanation to why the query is incorrect and re-prompts the LLM. 
The result is a new query which can then be passed back to the OBQC. 
Our approach gives us the opportunity to understand the capability of an LLM to repair a SPARQL query and thus further improve the accuracy.

\begin{figure}[hbtp]
\centering
\includegraphics[scale=0.4]{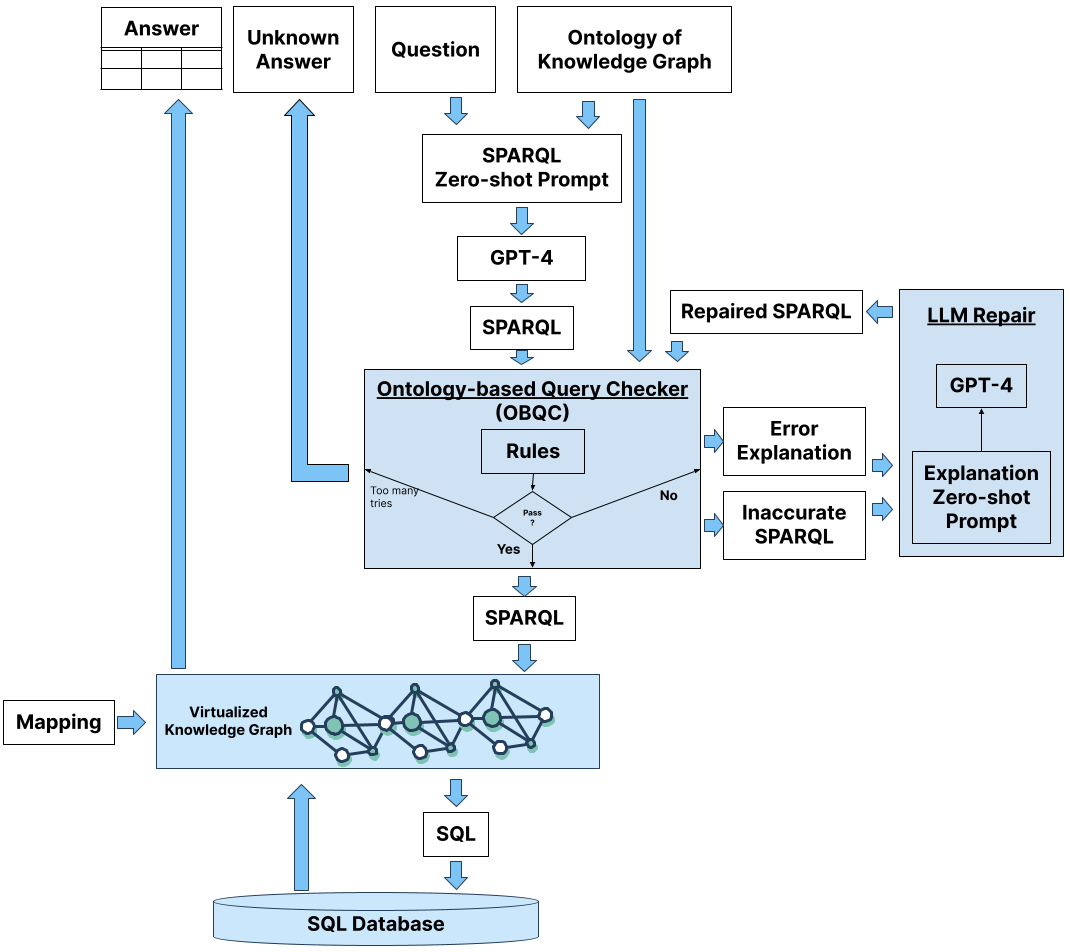}
\caption{Overview of our Ontology-based Query Checker and LLM Repair approach}
\label{fig:overview}
\end{figure}

Figure \ref{fig:overview} depicts the overview of our approach. 

We evaluated our approach using the chat with the data benchmark from our previous work\footnote{\url{https://github.com/datadotworld/cwd-benchmark-data}} \cite{OurPreviousWork}.
Our primary findings are the following: 
\begin{itemize}
    \item By grouping all the questions in the benchmark, the OBQC and LLM Repair increased the accuracy 72.55\%. If the repairs were not successful after three iterations, an unknown result was returned, which occurred 8\% of the time. The result is an error rate of 20\%. 
    \item Low complex questions on low complex schemas achieves an error rate of 10.46\%, which is now at levels deemed to be acceptable by users. 
    \item All questions on high complex schemas substantially increased the accuracy. 
    \item The rules related to the body of the query domain were invoked 70\% of the time while the rules related to the head of the query were invoked 30\% of the time. 
\end{itemize}

Comparing these results to our previous work \cite{OurPreviousWork}, we see a progression towards higher accuracy and reduced error rate. 

Previously, we showed that by using GPT-4 and zero-shot prompting, enterprise natural language questions over
enterprise SQL databases achieved 16.7\% accuracy.
This accuracy increased to 54.2\% when a Knowledge Graph
representation of the SQL database was used.
Thus an 3x accuracy improvement (37.5\%). 
By using the same zero-shot prompting on GPT-4 and adding \textbf{Ontology-Based Query Check} and \textbf{LLM Repair}, that new accuracy is 72.55\%.
That is over a 4x accuracy improvement compared to not using a knowledge graph at all. 

If we look at the error rate, we also see a substantial decrease. 
Without leveraging a knowledge graph, the error rate is 83.3\%.
The error rate decreased to 45.8\% when using a knowledge graph.
Given that we consider an unknown result in our approach because the \textbf{LLM Repair} is not able to repair and the \textbf{Ontology-Based Query Check} catches the error, the error rate has been further decreased to 19.44\%. 

The main conclusion is that we provide further strong evidence that investing in semantics, ontologies and knowledge graphs are prerequisites to achieve higher accuracy for LLM-powered question answering systems.

\section{Research Question and Hypothesis}
Our research goal is to understand how to increase the accuracy of Large Language Models powered question answering systems. 
By continuing to leverage knowledge graphs, our intuition is that the ontology can be used to detect errors in queries and an LLM be used to repair the query, thus improving the accuracy. 
Specifically, this intuition is based on the observation of our previous work \cite{OurPreviousWork}: the LLM generated a SPARQL query that produced inaccurate results due to the following: 
\begin{itemize}
    \item Incorrect Path: The generated query did not follow the correct path of the properties in the ontology. The generated path goes from A to C when the correct path is A to B to C. 
    \item Incorrect Direction: The generated query swaps the direction of a property. The generated direction is B to A, when the correct direction is A to B.
\end{itemize}

Therefore, we investigate the following research questions:

\noindent \textbf{RQ1}: To what extent can the accuracy increase by leveraging the ontology of a knowledge graph to detect errors of a SPARQL query and an LLM to repair the errors?
    
\noindent \textbf{RQ2}: What types of errors are most commonly presented in SPARQL queries generated by an LLM?

The hypothesis is the following: \textit{An ontology can increase the accuracy of an LLM powered question answering system that answers a natural language question over a knowledge graph.}

\section{Ontology-based Query Check}
Knowledge Graphs defined using the Semantic Web technology stack (specifically, RDF, RDFS, OWL and SPARQL) have been built on a rigorous logical foundation. 
The exact meaning of a statement (triple) in RDF is given in terms of predicate logic; the meaning of a model in RDFS or OWL is specified according to a logical foundation \cite{10.1145/3382097,DBLP:journals/csur/HoganBCdMGKGNNN21,DBLP:series/synthesis/2021Hogan}\footnote{For an introduction to knowledge graphs and semantic web technologies, we refer the reader to the textbooks ``Semantic Web for the Working Ontologist" and ``Knowledge Graphs" \url{https://kgbook.org/} }. 
The meaning of a SPARQL query is specified in terms of these logical foundations. 
A practical upshot of this theoretical framework is that it is possible to know exactly what constraints a model in RDFS or OWL places on the correctness of a SPARQL query, and these constraints can be described in an executable way in SPARQL. 
We leverage this foundation to create an ontology-based query check system for SPARQL queries. 
This service takes two inputs: a SPARQL query and an ontology, and returns one output; a list of sentences that describe ways in which the SPARQL query deviates from the specifications in the ontology. 

The check system relies on the declarative nature of SPARQL, the structure of Basic Graph Patterns and importantly, the ability to query the ontology via SPARQL itself. 
If the generated query deviates from the ontology, the approach outlines how.
The approach to achieve this is threefold:

First, a SPARQL query consists of a pattern to be matched against the data (specified after the keyword WHERE in the query); known as a Basic Graph Pattern (BGP) of the query. 
The process begins with extraction of BGPs from the generated SPARQL query, replacing variables with resources from a reserved namespace (prefixed with qq:). Some portions of the original query logic, including the SELECT clause, subquery structures, filters, UNIONs, OPTIONAL and NOT clauses, aren't considered since the focus is on examining the compatibility of the BGP with the ontology structure. We leave that for future work. 
However, note that violations of BGPs in an OPTIONAL or FILTER NOT EXISTS / MINUS context are not ignored as they can also provide vital insights into the query understanding.

Second, construction of a \textit{conjunctive graph}\footnote{using RDFLib nomenclature}, encapsulating two named graphs: \texttt{:query} and \texttt{:ontology}, representing the query BGP-turned-RDF and the ontology, respectively.

Third, application of ontology consistency rules guided by the formal logic of RDFS and OWL. These SPARQL-implemented rules examine the available \texttt{:query} and \texttt{:ontology} graphs and identify instances where the query diverges from the ontology.

\subsection{Body of a Query (WHERE Clause) Checks}
The following are the rules that check the Basic Graph Patterns of a SPARQL query (i.e. WHERE clause) matches the semantics of an ontology. 
In our current approach, we follow a subset of the RDF Schema (RDFS) semantics\footnote{\url{https://www.w3.org/TR/rdf11-schema/}}. 

\subsubsection{Domain Rule}
The domain rule (rdfs:domain in RDF Schema) is defined in English as follows: 
If the domain of a property p is a class C, then the subject of any triple using p as a predicate must be a member of class C.
The domain rule is formally defined as: 
\begin{verbatim}
IF 
    ?p rdfs:domain ?C . 
    ?s ?p ?o .  
THEN 
    ?s rdf:type ?C .
\end{verbatim}

The following SPARQL query is a representation of this domain rule which detects a violation: 
\begin{lstlisting}[language=SPARQL]
SELECT ?p ?domain ?s ?class WHERE {
    GRAPH :query{
      ?s ?p ?o .
      ?s a ?class .
    }
    GRAPH :ontology{
      ?p rdfs:domain ?domain.
      FILTER (ISIRI (?domain))
    }
    FILTER NOT EXISTS {
        ?class rdfs:subClassOf* ?domain .
    }
}
\end{lstlisting}

The following example shows how the domain rule is used to check a BGP against an ontology.
Suppose the LLM generated the following query: 

\begin{lstlisting}[language=SPARQL]
SELECT ?agent WHERE {
  ?agent :soldByAgent ?policy . 
  ?agent rdf:type :Agent 
}
\end{lstlisting}

This query has a BGP consisting of two-triples: 

\begin{verbatim}
 ?agent :soldByAgent ?policy . 
 ?agent rdf:type :Agent

\end{verbatim}

The BGP is turned into RDF graph by replacing the variables with resources from a reserved namespace (prefixed with qq:): 

\begin{verbatim}
  qq:agent :soldByAgent qq:policy . 
  qq:agent rdf:type :Agent
\end{verbatim}

Now, suppose the ontology includes the following definition of \texttt{:soldByAgent}: 

\begin{verbatim}
:soldByAgent 
  rdfs:domain :Policy ;
  rdfs:range :Agent .
\end{verbatim}

The conjunctive graph in nquads is the following: 

\begin{verbatim}
:query {
    qq:agent :soldByAgent qq:policy . 
    qq:agent rdf:type :Agent  .
 }
:ontology {
    :soldByAgent 
      rdfs:domain :Policy ;
      rdfs:range :Agent . 
}
\end{verbatim}

Let's look at this from the point of view of RDFS. 
The first clause, on the graph \texttt{:query}, finds the precondition for \texttt{rdfs:domain}; there is a triple with predicate \texttt(?p) whose subject is a member of some \texttt{class}.
The second clause searches the ontology for a relevant domain definition; that is, that same property \texttt{?p} has a specified domain.  This query ignores domain definitions that are not IRIs, which would typically include domains that are UNIONs or INTERSECTIONs of other classes. We have simplified the query for exposition in this paper, but it would be easy enough to extend the query to deal with other OWL constructs. We leave this as future work.
Finally, the FILTER clause of this query checks to make sure that the class specified in the  input query (\texttt{?class}) is not included in the domain (\texttt{?domain}). 
If all of these conditions in the evaluation query hold, then we have found a violation of the ontology in the query.  

Continuing our example, in the \texttt{:query} graph, we have a match for the first clause, with the binding 

\begin{verbatim}
?s -> qq:agent
?p -> :soldByAgent
?class -> :Agent
\end{verbatim}

The second clause searches \texttt{:ontology} for a triple matching

\begin{verbatim}
?soldByAgent rdfs:domain ?domain 
\end{verbatim}

this matches, with \texttt{?domain} bound to \texttt{:Policy} (which is indeed an IRI). 
Finally, we test whether 

\begin{verbatim}
:Agent rdfs:subClassOf* :Policy .
\end{verbatim}

Notice that the meaning of the * in SPARQL implies that this would succeed if :Agent were the same as :Policy.  
But in this case, they are not the same, and there is no such triple, so the FILTER NOT EXISTS condition succeeds, and the check comes up with a match, with the following bindings 

\begin{verbatim}
?p -> :soldByAgent
?domain -> :Policy
?s -> qq:agent
?class -> :Agent
\end{verbatim}

This information is not very understandable to a human, and might not be usable to an LLM.  
But it can be formatted into a meaningful sentence in English as follows: 

The property \texttt{:soldByAgent} has domain \texttt{:Policy}, but its subject \texttt{?agent} is a \texttt{:Agent}, which isn't a subclass of \texttt{:Policy}.

For each check rule, we provide a template that can create this explanation.  In this case, the template is:
\textit{The property \{p\} has domain \{dom\}, but its subject \{s\} is a \{class\}, which isn't a subclass of \{dom\}}

%There are five other rules that come from the definitions in RDFS.

\subsubsection{Range Rule}

The range rule (rdfs:range in RDF Schema) is defined in English as follows: 
If the range of a property p is a class C, then the object of any triple using p as a predicate must be a member of class C.
The range rule is formally defined as: 
\begin{verbatim}
IF 
    ?p rdfs:range ?C . 
    ?s ?p ?o .  
THEN 
    ?o rdf:type ?C .
\end{verbatim}

The following SPARQL query is a representation of this range rule which detects a violation:
\begin{lstlisting}[language=SPARQL]
SELECT ?p ?range ?s ?class  WHERE {
    GRAPH :query{
        ?s ?p ?o .
        ?o a ?class .
    }
    GRAPH :ontology{
      ?p rdfs:range ?range. FILTER (ISIRI (?range))
    }
    FILTER NOT EXISTS {?class rdfs:subClassOf* ?range .}
}

\end{lstlisting}
The message template is:
\textit{The property \{p\} has range \{range\}, but its object \{o\} is a \{class\}, which isn't a subclass of \{range\} }

Consider the following an example: an ontology that has one more step between a Claim and a Policy than many people might expect: 

\begin{verbatim}
in:against rdf:type owl:ObjectProperty ;
           rdfs:domain in:Claim ;
           rdfs:range in:PolicyCoverageDetail  .
in:hasPolicy rdf:type owl:ObjectProperty ;
             rdfs:domain in:PolicyCoverageDetail ;
             rdfs:range in:Policy  .
\end{verbatim}
that is, claims are not made directly against a policy, instead they are made against a PolicyCoverageDetail, which has a Policy.
Consider the following snippet of a LLM generated SPARQL query:

\begin{verbatim}
?policy rdf:type :Policy .
?claim rdf:type :Claim .    
?claim :against ?policy .
\end{verbatim}

We have observed this as a typical error that an LLM makes; it is relying on its background knowledge of policies and claims, rather than on the details of the ontology. 
How can we tell that this is an error? 

The analysis is very similar to the one for the domain rule; in this case, the range rule looks for the range of \texttt{:against}, which is \texttt{:PolicyCoverageDetail}.  But in the query, we see that the object of the predicate \texttt{:against} is the variable \texttt{?policy}, which in turn, has type \texttt{:Policy}. 

The template is filled out as 
\textit{The property :against has range :PolicyCoverageDetail, but its object ?policy is a :Policy, which isn't a subclass of :PolicyCoverageDetail.}

\subsubsection{Double Range Rule}

What if two triples make conflicting requirements on the range of a property? 

If the object of a first triple is the object of a second triple, then the range of the property of the first triple should be the same as the range of the property of the second triple. 

\begin{lstlisting}[language=SPARQL]
SELECT ?p ?rangep ?q ?rangeq WHERE {
    GRAPH :query {
      ?s1  ?p ?o .
      ?s2  ?q ?o .
    }
    GRAPH :ontology{
      ?p rdfs:range ?rangep.
      ?q rdfs:range ?rangeq.
      FILTER (ISIRI (?rangeq)) FILTER (ISIRI (?rangep))
    }
    
    FILTER NOT EXISTS {{
      { ?rangep rdfs:subClassOf* ?rangeq .}
      UNION 
      { ?rangeq rdfs:subClassOf* ?rangep .}
}}}
\end{lstlisting}

The explanation template: \textit{The property \{p\} has range \{rangep\}, and \{q\} has range \{rangeq\}, and these are incompatible. }

Consider the following snippet of an LLM generated SPARQL query

\begin{verbatim}
?claim :against ?policy . 
?policyCoverageDetail :hasPolicy ?policy .
\end{verbatim}

Now, suppose the ontology includes the following definitions: 

\begin{verbatim}
:against 
  rdfs:domain :Claim ;
  rdfs:range :PolicyCoverageDetail .

:hasPolicy 
  rdfs:domain :PolicyCoverageDetail ;
  rdfs:range :Policy .
\end{verbatim}

The LLM has tried to insert a PolicyCoverageDetail between the claim and the policy, but it didn't get it right.  
The range of \texttt{:against} and the range of \texttt{:hasPolicy} are \texttt{:PolicyCoverageDetail} and \texttt{:Policy} respectively.  
The check system outputs the following explanation: 
\textit{The property :against has range :PolicyCoverageDetail, and :hasPolicy has range :Policy, and these are incompatible.}

\subsubsection{Double Domain Rule}

What if two triples make conflicting requirements on the domain of a property? 

If the subject of a first triple is the subject of a second triple, then the domain of the property of the first triple should be the same as the domain of the property of the second triple.

\begin{lstlisting}[language=SPARQL]
SELECT ?p ?domp ?q ?domq WHERE {
    GRAPH :query{
      ?s ?p ?o1 .
      ?s ?q ?o2 .
    }
    GRAPH :ontology{
      ?p rdfs:domain ?domp.
      ?q rdfs:domain ?domq.
      FILTER (ISIRI (?domq)) FILTER (ISIRI (?domp))
    }
    FILTER NOT EXISTS {{
      { ?domp rdfs:subClassOf* ?domq .}
      UNION 
      { ?domq rdfs:subClassOf* ?domp .}
}}}
\end{lstlisting}

The template: \textit{The property \{p\} has domain \{domp\}, and \{q\} has domain \{domq\}, and these are incompatible. }

\subsubsection{Domain Range Rule}

If the object of a first triple is the subject of a second triple, then the range of the property of the first triple should be the same as the domain of the property of the second triple. 

\begin{lstlisting}[language=SPARQL]
SELECT ?p ?rangep ?q ?domq
WHERE {
    GRAPH :query 
    {
      ?s ?p ?o  .
       ?o  ?q ?o2 .
         
    }
    GRAPH :ontology 
    {
      ?p rdfs:range ?rangep .
      ?q rdfs:domain ?domq.
      FILTER (ISIRI (?domq))
      FILTER (ISIRI (?rangep))
      
    }
    
    FILTER NOT EXISTS 
  {
    {
      { ?rangep rdfs:subClassOf* ?domq .}
      UNION 
      { ?domq rdfs:subClassOf* ?rangep .}
      }
    }
}
\end{lstlisting}

The template: \textit{The property \{p\} has range \{rangep\}, and \{q\} has domain \{domq\}, and these are incompatible with the query. }

\subsubsection{Incorrect Property}

All the properties in the query need to exist in the ontology. 

\begin{lstlisting}[language=SPARQL]
SELECT ?p WHERE {
    GRAPH :query { 
    ?s ?p ?o .
    FILTER NOT EXISTS {
      VALUES ?ns {
	  "http://www.w3.org/1999/02/22-rdf-syntax-ns#"
        "http://www.w3.org/2002/07/owl#" 
        "http://www.w3.org/2000/01/rdf-schema#"
	  "http://www.w3.org/2004/02/skos/core#"
      }
      FILTER(STRSTARTS(STR(?p), ?ns))
    }}
  FILTER NOT EXISTS {
    GRAPH :ontology {
      ?p a ?type 
}}}
\end{lstlisting}

The template is: \textit{The property \{p\} isn't defined in the ontology.  Please only use properties from the ontology, or from a standard source like rdf:, rdfs:, owl:, or skos:
}
\subsection{Head of a Query (SELECT Clause) Checks}
The following are the rules that make checks to the head of a SPARQL query (i.e. SELECT clause). 
A common error for an LLM is to include extra values in the SELECT clause or to leave some out. 
These errors have nothing to do with the ontology. 
However, a very common error is to include a variable in the SELECT clause that will be bound to an IRI (an identifier). 
It is not usually good practice to return IRIs as values for a query that will be viewed by a business user; there is no guarantee that the IRI will be meaningful (and in fact, a good reason to expect that it will not be meaningful). 
There are some constructs in the ontology that can predict whether a value will be an IRI; for example, the RDF specification states that the subject of any triple is an IRI; if a query selects a variable that is the subject of a basic triple pattern, we can tell that it is an IRI.  
Similarly, if a predicate has a specified range which is a class, then the object of that triple can be inferred to be an IRI.  
We have extended this method to identify variables in the SELECT clause and indicate them in the conjunctive graph, so that we can detect these situations as well. 

\noindent\textbf{IRI Output Rule}

\begin{lstlisting}[language=SPARQL]
SELECT  ?varname WHERE {
    GRAPH :query {
      ?s ?p ?o .
      ?o a qq:Variable .
      BIND  (REPLACE(STR(?o), "^.*[/#]", "") as ?varname)
    }
    GRAPH :query {
        ?p rdfs:range ?range.
        FILTER (ISIRI(?range))
}}
\end{lstlisting}
The template is: \textit{Your selected variable \{varname\} is an IRI; your output should be something human readable, an ID or a label. }

\noindent\textbf{Subject Output Rule}

\begin{lstlisting}[language=SPARQL]
SELECT  ?varname WHERE {
    GRAPH :query {
      ?s ?p ?o .
      BIND  (REPLACE(STR(?s), "^.*[/#]", "") as ?varname)
      ?s a qq:Variable . 
}}
\end{lstlisting}
The template is: \textit{Your selected variable ?\{varname\} is an IRI (the subject of a triple is always an IRI). Your output should be something human readable, an ID or a label. }

\subsection{Future work}
This approach takes advantage of the power of an ontology to describe data and metadata, but it has some known limitations. 

Increased ontology expressivity modeled in OWL where domains or ranges represented as union or other logical combinations of classes may break this implementation. 
We mentioned this above, when we noticed the filter that insisted that the domain must be an IRI.  This limitation can be repaired by using more complex SPARQL queries, but trying to implement the entire logical definition of OWL in SPARQL is not only difficult, in general, it is not possible. 
However, for most ontologies that are actually used in industry, simple rules like the ones outlined here are sufficient. 
 
The hierarchical checks made assume that classes not explicitly defined to be subclasses of one another are disjoint - an assumption not guaranteed by OWL's Open World Assumption. 
It is common practice, even in very well-governed OWL ontologies, not to include all of the disjointness axioms that are known to be true, effectively making something of a closed-world assumption in many parts of an OWL model. 

Despite these constraints, the check system provides an effective way to ensure the generated SPARQL queries adhere to the underlying ontology, as we show in the results.

\section{LLM Repair: Repairing SPARQL queries with an LLM}

Given that we have a facility that can take a question and an ontology and generate a query, and another facility that can take a query and an ontology and generate critiques of that query, how do we use this to get a more reliable question answering system? 
We need one more component, one that can take those critiques and use them to improve the query. That is the LLM Repair.  

The input to the LLM Repair is a prompt that consists of 1) the list of issues for which the query is incorrect which is the output of the OBQC and 2) the incorrect SPARQL query:

\noindent\textbf{Explanation Zero-shot Prompt}
\begin{verbatim}
We have a query {query} with some issues outlined here {issues}
Please re-write it.    
\end{verbatim}

The output is a new LLM generated SPARQL query, which is passed again to the OBQC.
This cycle repeats until the check pass, or an upper limit of cycles is reached. 
In our experiments, the limit is 3. 
In this latter case, the query generation is said to be unknown; there is no point in sending a query that is known to be faulty to the database. 

It is instructive to note that the LLM repair focuses on that task; we do not repeat the question nor the ontology to the LLM. 
The ontology input was taken into consideration by the OBQC, and the question is reflected in the query so far. 

It is also noteworthy that there are two possible outcomes; we can achieve a query that we have considerable confidence in (because it matches the semantics of the ontology), or we fail to create such a query. 
In the latter case, we are aware of the failure of the system. 
In contrast to a pure LLM-based system which is prone to \textit{hallucinations}, when we get the wrong answer, we know it, and can report that to the user. 

\section{Experimental Setup}
The experiments reported here use the same data as the Chat with the Data benchmark from our previous work, which is available on github\footnote{\url{https://github.com/datadotworld/cwd-benchmark-data}}.
As a reminder, that benchmark consists of 
\begin{enumerate}
    \item Enterprise SQL Schema based on the OMG Property and Casualty Data Model 
    \item Enterprise question-answer that fall on a combination of two spectrums:
    \begin{itemize}
        \item Low to High question complexity pertaining to business reporting use cases to metrics and Key Performance Indicators (KPIs) questions
        \item  Low to High schema complexity which requires a smaller number of tables
to larger number of tables to answer the question
    \end{itemize}
These two spectrums form a quadrant in which questions
can be classified: Low Question/Low Schema, HighQuestion/Low Schema, Low Question/High Schema, and
High Question/High Schema. 
\item Context Layer: An OWL ontology describing the Business Concepts, Attributes, and Relationships of the insurance
domain, an R2RML mappings from the SQL schema to the OWL ontology. The ontology and mappings can be used to
create a Knowledge Graph representation of the SQL database.
\end{enumerate}

The question answering system we evaluated was a SPARQL zero-shot prompt to GPT-4, that is instructed to generate a query, which is executed over a virtualized knowledge graph in data.world that is mapped to a relational database. This is the same setup as in our previous work \cite{OurPreviousWork}:

\noindent\textbf{SPARQL Zero-shot Prompt}

\begin{verbatim}
Given the OWL model described in the following TTL file: 
{INSERT OWL ontology}
Write a SPARQL query that answers the question.   
Do not explain the query.  Return just the query, 
so it can be run verbatim from your response.
Here's the question: {INSERT QUESTION}
\end{verbatim}

% This experiment takes advantage of the query checking capabilities outlined in the previous section. 
% An ontology can do more than  advise an LLM about how to write a query; 
% it can also evaluate whether a query ``makes sense", i.e., the semantics of the query matches the ontology. 
% We use this capablitiy in two ways: first, to find issues with a query and repair 
% it as described above. But second, and perhaps even more importantly, this allows us to detect a query that we know for certain cannot succeed. 
In addition to Accurate queries (that get the right answer), Inaccurate queries (that get the wrong answer), we have {\it unknown} queries, queries that we know are incorrect. 
This is an important category, as it contrasts with the well-known and oft-maligned "\textit{hallucination}" behavior of LLMs, where it provides an incorrect answer. 
In this setup, we also have answers that we know to be incorrect. This third kind of
result, {\it unknown}, plays into all of the metrics we track in the results. 

As we summarize the results, we follow the metric of Execution Accuracy (EA) from the Spider benchmark \cite{data-spider}.
An execution is accurate if the result of the query matches the answer for the query. 
Note that the order or the labels of the columns are not taken in account for accuracy. 

We report the following metrics for a SPARQL query generated by an LLM: 

\begin{itemize}
    \item Execution Accuracy First Time: if the OBQC returns true the first time which results in an accurate execution.
    \item Execution Accuracy with Repairs: if the OBQC returns false the first time and the LLM Repair results in an accurate execution. 
    \item Execution Unknown with Repairs: if the OBQC returns false the first time and the LLM Repair is unable to repair after three attempts. 
\end{itemize}

\section{Results}

%In order to provide an answer to the overall research question, \textit{to what extent can the ontology of a Knowledge Graph be used to improve the accuracy of a Large Language Model (LLM) based question answering system?}, 
We present the results addressing the two specific research questions: 

\subsection{Results for RQ1: To what extent can the accuracy increase by leveraging the ontology of a knowledge graph to detect errors of a SPARQL query and an LLM to repair the errors?}

By grouping all the questions in the benchmark, the Average Overall Execution Accuracy with Repairs is 72.55\%. 
This is an increase of 29.67\% based on the Average Overall Execution Accuracy First Time which is 42.88\%.
The Average Overall Execution Unknown with Repairs is 8\% which implies that the LLM Repair is usually able to repair the queries and is still able to identify when queries can not be repaired.
By combining the Average Overall Execution Accuracy with Repairs and Average Overall Execution Unknown with Repairs, the error rate is 19.44\%. 

Based on the results shown in \ref{table:Results}, we observe that the Ontology-based Query Check and LLM Repair favorably increased the accuracy and reduced the error rate in two areas: 
\begin{itemize}
    \item Questions on High Complex Schema: the Ontology-based Query Check and LLM Repair positively impacts the accuracy of all types of questions that are on a high schema complexity.
    \item Combining Accuracy and Unknowns: ``I don't know" is a valid answer and arguable a better answer than an inaccurate answer. By combining accuracy and unknown, the error rate reduces, notably making a bigger impact in Low Question/Low Schema.

\end{itemize}

\begin{table}[h!]
\centering
\begin{tabular}{ | >{\raggedright\arraybackslash}m{2.7cm}|>{\raggedleft\arraybackslash} m{2.05cm}| >{\raggedleft\arraybackslash}m{2.15cm} | >{\raggedleft\arraybackslash}m{2.15cm} | >{\raggedleft\arraybackslash} m{2.15cm} |>{\raggedleft\arraybackslash}m{2.15cm} | >{\raggedleft\arraybackslash} m{2.15cm} |} 
\hline
  & \textbf{Average Overall Execution Accuracy First Time} & \textbf{Average Overall Execution Accuracy with Repairs} & \textbf{Average Overall Execution Unknown with Repairs}& \textbf{Average Overall Execution Accuracy + Unknown with Repairs} & \textbf{Error Rate}\\ 
  \hline
  \textbf{All Questions}          & 42.88\% & 72.55\%    & 8\%     & 80.56\%  & 19.44\% \\ 
  \hline
  \textbf{Low Question / Low Schema}& 51.19\% & 76.67\%    & 12.87\%      & 89.54\%  & 10.46\% \\ 
  \hline
  \textbf{High Question / Low Schema}& 69.76\% & 75.10\%    & 6.02\%    & 81.12\%  & 18.88\% \\  
  \hline
  \textbf{Low Question / High Schema}& 17.20\% & 76.33\%    & 3.45\%    & 79.79\%  & 20.21\% \\   
  \hline
  \textbf{High Question / High Schema}& 28.17\% & 60.62\%  & 8.40\%    & 69.03\%  & 30.97\% \\ 
  \hline
\end{tabular}
\caption{Average Overall Execution Accuracy (AOEA) of Overall and Quadrant Results}
\label{table:Results}
\end{table}

We observe the following details for each quadrant: 
\begin{itemize}
    \item Low Question/Low Schema: the Ontology-based Query Check and LLM Repair increased the accuracy by 25.48\%. The Execution Unknown with Repairs was the highest in this quadrant, 12.87\%. Combined, this implies that the error rate is 10.46\%, the lowest of all the quadrants.  
    \item High Question/Low Schema: the Ontology-based Query Check and LLM Repair increased the accuracy by 5.34\% which was the lowest increase of the quadrants. It is not clear why. The final error rate is 18.88\%. 
    \item Low Question/High Schema: the Ontology-based Query Check and LLM Repair had a substantial impact by increasing the accuracy by 59.13\% with an error rate of 20.21\%.
    \item High Question/High Schema: the Ontology-based Query Check and LLM Repair had a meaningful impact by increasing the accuracy by 32.45\% with an error rate of 30.97\%.
\end{itemize}

\begin{figure}[hbtp]
\centering
\includegraphics[scale=0.5]{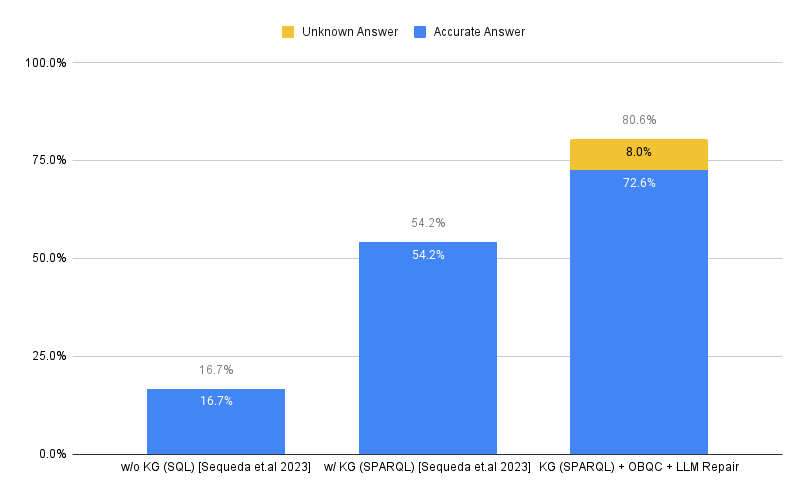}
\caption{Average Overall Execution Accuracy (AOEA) of SPARQL and SQL for all the questions in the benchmark from \cite{OurPreviousWork} compared to OBQC and LLM Repair}
\label{fig:OverallResultsWithPrevious}
\end{figure}

\begin{figure}[hbtp]
\centering
\includegraphics[scale=0.35]{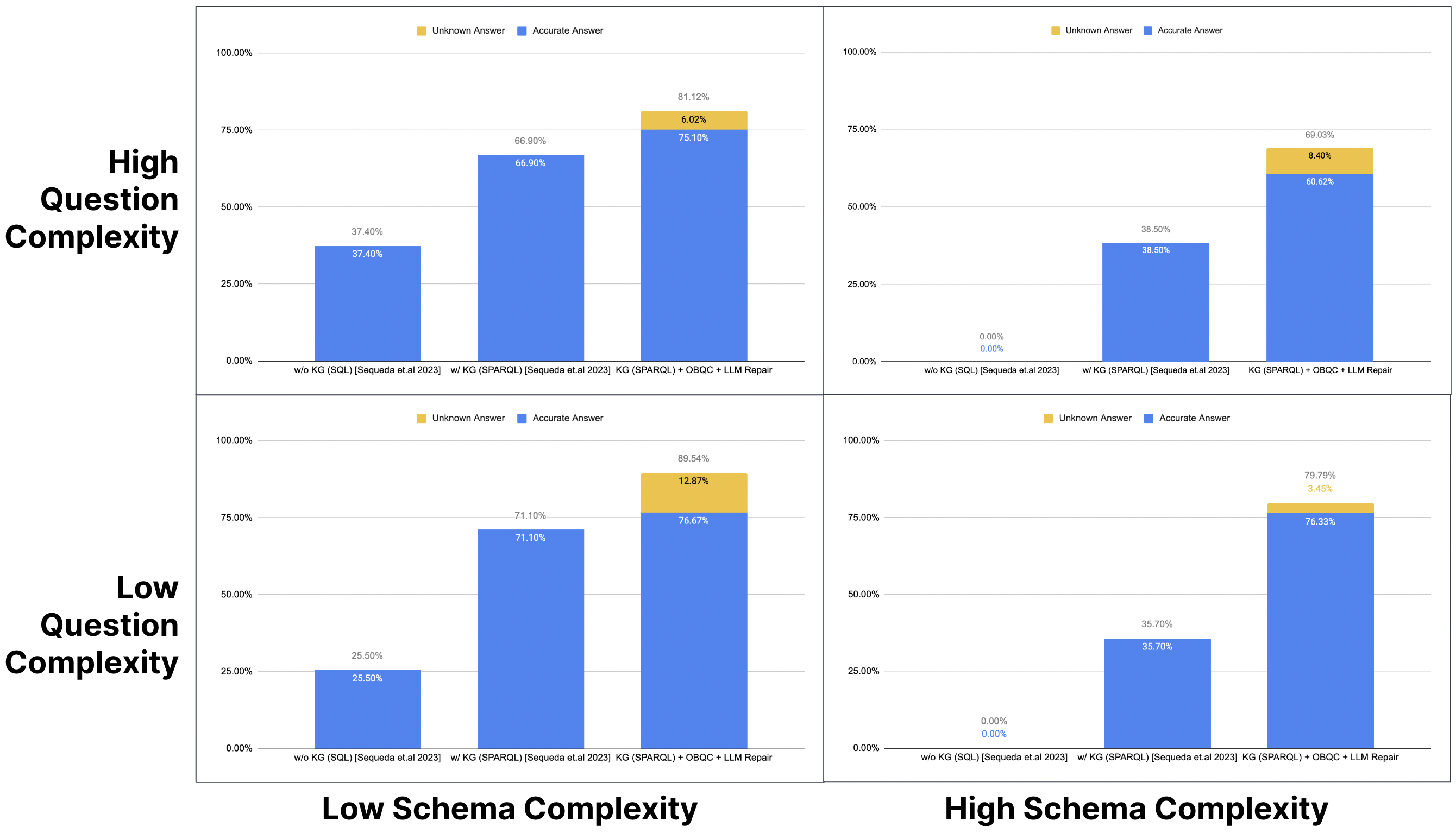}
\caption{Average Overall Execution Accuracy (AOEA) of SPARQL and SQL for all questions in each quadrant from \cite{OurPreviousWork} compared to OBQC and LLM Repair}
\label{fig:QuadrantResultsWithPrevious}
\end{figure}

Comparing to the results of our previous work\cite{OurPreviousWork}, the accuracy increase is notable. The Average Overall Execution Accuracy of the same zero-shot Text-to-SPARQL prompt on a Knowledge Graph representation of the SQL database, reported in our previous work, was 54.2\%, indicating an error rate of 45.8\%. 
With our Ontology-based Query Check and LLM Repair, the error rate is reduced to 19.44\%. 
Figure \ref{fig:OverallResultsWithPrevious} depicts these results for all the questions in the benchmark.
Figure \ref{fig:QuadrantResultsWithPrevious} depicts these results for all questions in each quadrant.

These results are evidence that supports our hypothesis, ontologies can improve the accuracy of LLM powered question answering systems.

A follow up question to understand the \textit{extent} is the following: 
\textit{How much of the possible execution accuracy improvement was achieved?}
In other words, given the number of times the system achieved an accurate answer on the first time, and the total number of runs, we know how much is left for improvement. Therefore, how much of that improvement was achieved? 
For example, given a total of 10 runs, where 2 of them were accurate on the first try achieving a First Time Execution Accuracy of 20\%, means that there are 8 runs left where the OBQC and LLM Repair to repair a query in order to achieve 100\% Execution Accuracy with Repairs. Let's say that the system is able to accurately repair 4 times, therefore the Execution Accuracy with Repairs is 60\%. 
The achievable improvement is 50\% because the accurate repair occurred 4 times out of the 8 possible times. 
Achievable Improvement is calculated as (Number of Accurately Repaired Queries) / (Total Number of runs - Number of First Time Executed Accurate queries).
The results of achievable improvement are shown in Table \ref{table:AchievableImprovementResults}

\begin{table}[h!]
\centering
\begin{tabular}{ | m{5cm} | m{5cm} |} 
\hline
  & \textbf{Average Achievable Improvement}\\ 
  \hline
  \textbf{All Questions}   & 55.57\% \\ 
  \hline
  \textbf{Low Question/Low Schema}& 49.30\% \\ 
  \hline
  \textbf{High Question/Low Schema}& 40.45\% \\ 
  \hline
  \textbf{Low Question/High Schema}& 72.23\% \\  
  \hline
  \textbf{High Question/High Schema}& 57.70\% \\ 
  \hline
\end{tabular}
\caption{Average Achievable Improvement of Overall and Quadrant Results}
\label{table:AchievableImprovementResults}
\end{table}

The results indicate that the OBQC is able to successfully repair queries half of the time. 
Thus, we are halfway there and there is still room for improvement. 
Recall that OBQC mainly checks the body of the query and just two checks in the SELECT clause determine if IRI identifiers are returned. 
We postulate that a set of inaccurate queries are due to the overlap type of partial accurate queries: the columns returned by the query are correct, however, they are a subset of the accurate answer.
Therefore possible repair rules can be defined to check the head of the query.
Additionally, this may indicate the need for more expressive ontologies. 

\subsection{Results for RQ2: What types of errors are most commonly presented in SPARQL queries generated by an LLM?}

In our experiments, we kept count of the number of times a rule was invoked by the OBQC.
Table \ref{table:RuleUsage} presents the percentage of usage of each rule in the OBQC. 

Notably, 70\% of the repairs were done by the Ontological checks while 30\% of the repairs were done by the SELECT Clause checks. 

The rules exclusively related to domain were invoked 42.16\% of the time and surprisingly, rules exclusively related to range were invoked less than 1\% of the time. 
The Domain Range rule contributed to 22.78\% of the repairs.

\begin{table}[h!]
\centering
\begin{tabular}{ | m{3cm} | m{2cm} |} 
\hline
  \textbf{Rule} & \textbf{Usage}\\ 
  \hline
  Double Domain   & 37.47\% \\ 
  \hline
  Domain Range   & 22.78\% \\ 
  \hline
  IRI Output& 18.40\% \\ 
  \hline
  Subject Output& 11.91\% \\  
  \hline
  Domain& 4.69\% \\ 
  \hline
  Incorrect Property& 4.26\% \\ 
  \hline
  Range& 0.43\% \\ 
  \hline
  Double Range& 0.06\% \\ 
  \hline
\end{tabular}
\caption{Usage of the rules in the Ontology-based Query Check}
\label{table:RuleUsage}
\end{table}

A surprising result is that the domain related rules had the largest impact in repairs. 
These results may shine some light on what is happening underneath the hood inside an LLM, namely the relationship between english language and triples of a graph. 
English is written and read from left to right. 
The domain of a property has a relationship to the left side of a triple.
If the LLM writes a query which is wrong, it would most probably get it wrong at the beginning of a sentence/triple. 
This may be an explanation on why the domain related rules were the most impactful.

\section{Summary, Impact and Conclusion}

\noindent\textbf{Summary:} We present an approach to increase the accuracy of LLM-powered question answering systems that leverages the ontology of a knowledge graph. 
Our approach comes in two parts:
1) \textbf{Ontology-based Query Check} checks a SPARQL query generated by an LLM against the semantics of an ontology to identify and explain erroneous queries, and
2) \textbf{LLM Repair} takes the explanation and the erroneous SPARQL query and prompts the LLM to repair. 
This cycle repeats until the check passes, or an upper limit of cycles is reached.
Our experimental results provide evidence that our approach increases the accuracy of Large Language Model (LLM) question-answering systems to 72\%, including an additional 8\% of unknown, thus with an overall error rate of 20\%.

\noindent\textbf{Impact:} It is clear that customers desire to have a \textit{chat with your data} experience as if they are talking to an expert that understands their business and is able to provide answers that are accurate, explainable and governed. 
The Ontology-based Query Check and LLM Repair are just two components of larger agent framework that underpins data.world's AI Context Engine\footnote{\url{https://data.world/ai/}} which is an API-driven application that connects our customer's data and metadata with Generative AI to power trusted conversations with structured data. 
These two components are the results of a series of hackathons that the authors have conducted with various data.world customers since October 2023. 
The data.world AI Context Engine is in production use by a variety of data.world customers. 
The research and results presented in this paper are based on a co-innovation approach between the data.world AI Lab and data.world's customers.  

\noindent\textbf{Conclusion:} 
Our findings provide evidence that the use of ontologies can further increase the accuracy of LLM question-answering systems, thus supporting organisation's trust in the GenAI solutions they deploy. 
With these additions, we have moved toward a system that is more reliable, less prone to common errors, and ultimately more practical for usage in real-world applications.
These results support the main conclusion of this research: investment in metadata, semantics, ontologies and Knowledge Graph are preconditions to achieve higher accuracy for LLM powered question answering systems. 

% {\small
% \paragraph{Acknowledgement}
% We thank all our colleagues at data.world who supported our work. 
% We are also extremley thankful for the early feedback we have received on this work from colleagues across industry and academia in the fields of AI, Databases and Knowledge Graphs: 
% Albert Merono Penuela, 
% Dan Bennett, 
% Deborah McGuinness, 
% Diego Collarana, 
% Elena Simperl, 
% Ethan Mollick, 
% Gary George, 
% Gary Marcus, 
% George Fletcher, 
% Luke Slotwinksi, 
% Malcolm Chisholm, 
% Mark Kitson, 
% Michael Murray, 
% Mike Dillinger, 
% Mohammed Aaser,
% Olaf Hartig, 
% Omar Khawaja, 
% Ora Lassila, 
% Oscar Corcho, 
% Patrick van de Belt, 
% Paul Groth, 
% Peter Lawerence, 
% Rachel Wood, 
% Steve Gustafan, 
% Tony Seale, 
% Vip Parmar.
% }

\bibliographystyle{acm}
%\bibliography{references}  %%% Uncomment this line and comment out the ``thebibliography'' section below to use the external .bib file (using bibtex) .

\begin{thebibliography}{10}

\bibitem{10.1145/1514894.1514899}
{\sc Afrati, F.~N., and Kolaitis, P.~G.}
\newblock Repair checking in inconsistent databases: algorithms and complexity.
\newblock In {\em Proceedings of the 12th International Conference on Database Theory\/} (New York, NY, USA, 2009), ICDT '09, Association for Computing Machinery, p.~31–41.

\bibitem{10.1145/3382097}
{\sc Allemang, D., Hendler, J., and Gandon, F.}
\newblock {\em Semantic Web for the Working Ontologist: Effective Modeling for Linked Data, RDFS, and OWL}, 3~ed., vol.~33.
\newblock Association for Computing Machinery, New York, NY, USA, 2020.

\bibitem{bouzenia2024repairagent}
{\sc Bouzenia, I., Devanbu, P., and Pradel, M.}
\newblock Repairagent: An autonomous, llm-based agent for program repair, 2024.

\bibitem{data-atis-original}
{\sc Deborah A.~Dahl, Madeleine~Bates, M. B. W. F. K. H.-S. D. P. C. P. A.~R., and Shriber, E.}
\newblock {Expanding the scope of the ATIS task: The ATIS-3 corpus}.
\newblock {\em Proceedings of the workshop on Human Language Technology\/} (1994), 43--48.

\bibitem{10.1145/3318162}
{\sc Goues, C.~L., Pradel, M., and Roychoudhury, A.}
\newblock Automated program repair.
\newblock {\em Commun. ACM 62}, 12 (nov 2019), 56–65.

\bibitem{10.1145/1460690.1460714}
{\sc Green, B.~F., Wolf, A.~K., Chomsky, C., and Laughery, K.}
\newblock Baseball: An automatic question-answerer.
\newblock In {\em Papers Presented at the May 9-11, 1961, Western Joint IRE-AIEE-ACM Computer Conference\/} (New York, NY, USA, 1961), IRE-AIEE-ACM '61 (Western), Association for Computing Machinery, p.~219–224.

\bibitem{10.1145/800186.810578}
{\sc Green, C.~C., and Raphael, B.}
\newblock The use of theorem-proving techniques in question-answering systems.
\newblock In {\em Proceedings of the 1968 23rd ACM National Conference\/} (New York, NY, USA, 1968), ACM '68, Association for Computing Machinery, p.~169–181.

\bibitem{10.1145/320251.320253}
{\sc Hendrix, G.~G., Sacerdoti, E.~D., Sagalowicz, D., and Slocum, J.}
\newblock Developing a natural language interface to complex data.
\newblock {\em ACM Trans. Database Syst. 3}, 2 (jun 1978), 105–147.

\bibitem{DBLP:series/synthesis/2021Hogan}
{\sc Hogan, A., Blomqvist, E., Cochez, M., d'Amato, C., de~Melo, G., Gutierrez, C., Kirrane, S., Gayo, J. E.~L., Navigli, R., Neumaier, S., Ngomo, A.~N., Polleres, A., Rashid, S.~M., Rula, A., Schmelzeisen, L., Sequeda, J., Staab, S., and Zimmermann, A.}
\newblock {\em Knowledge Graphs}.
\newblock Synthesis Lectures on Data, Semantics, and Knowledge. Morgan {\&} Claypool Publishers, 2021.

\bibitem{DBLP:journals/csur/HoganBCdMGKGNNN21}
{\sc Hogan, A., Blomqvist, E., Cochez, M., d'Amato, C., de~Melo, G., Gutierrez, C., Kirrane, S., Gayo, J. E.~L., Navigli, R., Neumaier, S., Ngomo, A.~N., Polleres, A., Rashid, S.~M., Rula, A., Schmelzeisen, L., Sequeda, J.~F., Staab, S., and Zimmermann, A.}
\newblock Knowledge graphs.
\newblock {\em {ACM} Comput. Surv. 54}, 4 (2022), 71:1--71:37.

\bibitem{10.1145/3611643.3613892}
{\sc Jin, M., Shahriar, S., Tufano, M., Shi, X., Lu, S., Sundaresan, N., and Svyatkovskiy, A.}
\newblock Inferfix: End-to-end program repair with llms.
\newblock In {\em Proceedings of the 31st ACM Joint European Software Engineering Conference and Symposium on the Foundations of Software Engineering\/} (New York, NY, USA, 2023), ESEC/FSE 2023, Association for Computing Machinery, p.~1646–1656.

\bibitem{OurPreviousWork}
{\sc Sequeda, J., Allemang, D., and Jacob, B.}
\newblock A benchmark to understand the role of knowledge graphs on large language model's accuracy for question answering on enterprise sql databases, 2023.

\bibitem{data-restaurants-logic}
{\sc Tang, L.~R., and Mooney, R.~J.}
\newblock Automated construction of database interfaces: Intergrating statistical and relational learning for semantic parsing.
\newblock In {\em 2000 Joint SIGDAT Conference on Empirical Methods in Natural Language Processing and Very Large Corpora\/} (2000), pp.~133--141.

\bibitem{data-spider}
{\sc Tao~Yu, Rui~Zhang, K. Y. M. Y. D. W. Z. L. J. M. I. L. Q. Y. S. R. Z.~Z., and Radev, D.}
\newblock Spider: A large-scale human-labeled dataset for complex and cross-domain semantic parsing and text-to-sql task.
\newblock In {\em Proceedings of the 2018 Conference on Empirical Methods in Natural Language Processing\/} (2018), pp.~3911--3921.

\bibitem{10.1145/355598.362773}
{\sc Woods, W.~A.}
\newblock Transition network grammars for natural language analysis.
\newblock {\em Commun. ACM 13}, 10 (oct 1970), 591–606.

\bibitem{yan2024corrective}
{\sc Yan, S.-Q., Gu, J.-C., Zhu, Y., and Ling, Z.-H.}
\newblock Corrective retrieval augmented generation, 2024.

\bibitem{data-geography-original}
{\sc Zelle, J.~M., and Mooney, R.~J.}
\newblock Learning to parse database queries using inductive logic programming.
\newblock In {\em Proceedings of the Thirteenth National Conference on Artificial Intelligence - Volume 2\/} (1996), pp.~1050--1055.

\bibitem{10.1145/3631974}
{\sc Zhang, Q., Fang, C., Ma, Y., Sun, W., and Chen, Z.}
\newblock A survey of learning-based automated program repair.
\newblock {\em ACM Trans. Softw. Eng. Methodol. 33}, 2 (dec 2023).

\end{thebibliography}

% \newpage
% \section{Appendix}

\end{document}